# Hyperspectral Remote Sensing Image Classification Based on Multi-scale Cross Graphic Convolution

Yunsong Zhao[1,+] · Yin Li[1,+] · Zhihan Chen[2] · Tianchong Qiu[3] · Guojin Liu[1,*]

**Abstract** The mining and utilization of features directly affect the classification performance of models used in the classification and recognition of hyperspectral remote sensing images. Traditional models usually conduct feature mining from a single perspective, with the features mined being limited and the internal relationships between them being ignored. Consequently, useful features are lost and classification results are unsatisfactory. To fully mine and utilize image features, a new multi-scale feature-mining learning algorithm (MGRNet) is proposed. The model uses principal component analysis to reduce the dimensionality of the original hyperspectral image (HSI) to retain 99.99% of its semantic information and extract dimensionality reduction features. Using a multi-scale convolution algorithm, the input dimensionality reduction features were mined to obtain shallow features, which then served as inputs into a multi-scale graph convolution algorithm to construct the internal relationships between eigenvalues at different scales. We then carried out cross fusion of multi-scale information obtained by graph convolution, before inputting the new information obtained into the residual network algorithm for deep feature mining. Finally, a flexible maximum transfer function classifier was used to predict the final features and complete the classification. Experiments on three common hyperspectral datasets showed the MGRNet algorithm proposed in this paper to be superior to traditional methods in recognition accuracy.

## 1 Introduction

Today, hyperspectral remote sensing technology has become a very important part of remote sensing technology. With the development of hyperspectral image (HSI) fields, HSI classification has been greatly improved in terms of spatial and spectral resolution. Moreover, in the field of remote sensing—especially in HSI processing—deep learning methods have been widely used, including methods such as target detection, image segmentation, and image classification. Xuefeng Liu *et al.* [1] used CNN and deep belief network (DBN) deep learning models and traditional SVM methods to simulate real Arivis hyperspectral data. Jakub Nalepa *et al.* [2] proposed a deep learning technique for unsupervised HSI segmentation in the field of image segmentation. The problem of a limited real hyperspectral set on the ground can be solved by providing end-to-end unsupervised HIS subdivision. Anyong Qin *et al.* [3] proposed a framework of $S^2$ graph convolutional network (GCNs) model and verified a significant improvement of classification accuracy in the traditional HSI dataset.

Researches based on hyperspectral remote sensing images have shown that the feature-mining ability of hyperspectral remote sensing images directly affects the classification accuracy of a model. Consequently, on the basis of the existing methods of feature selection and feature integration, improving the feature-mining ability of HSIs has become the focus of hyperspectral remote sensing image processing. Zesong Wang *et al.* [4] proposed a sequential joint deep learning algorithm. The model uses a multi-scale convolution algorithm to solve the problem that a single convolution kernel is insufficient to mine information—that is, much useful information is potentially lost. Their algorithm greatly improves the feature-mining ability of the model but does not establish the relationships between eigenvalues, which means that some key information is still lost.

Although greater achievements have been made by CNNs, the traditional CNNs cannot fully capture the spectral spatial information of different local regions. In order to extract the spatial and spectral information of HSIs, Qichao Liu *et al.* [5] used graph convolution to better simulate the spatial context structure of HSIs by coding HSIs into graphics and making use of the correlation between adjacent land cover, which successfully established the relationships between features and effectively improved the classification accuracy. Sheng Wan *et al.* [6] proposed a multi-scale graph convolution algorithm, which used graphs of different scales for convolution, establishing a more comprehensive relationship between features.

However, when processing hyperspectral remote sensing

✉ Guojin Liu
liuguojin@cqu.edu.cn

[1] School of Microelectronics and Communication Engineering, Chongqing University, Chongqing 401331, China

[2] School of Automation, Chongqing University, Chongqing 401331, China

[3] Glasgow College, UESTC, University of Electronic Science and Technology of China, Chengdu 611731, China

* Correspondence

+ The first two authors have equal contribution

images based on multi-scale image convolution, researchers often deal with feature information at different scales while ignoring the internal relationships between them (at different scales). To solve these problems, we propose a new multi-scale graph convolution combined with a residual network fusion algorithm. Specifically, shallow features are mined using multi-scale convolution, and then the relationships between eigenvalues at different scales is established using multi-scale graph convolution. Next, we continue pairwise cross fusion on the output results of the convolution (at different scales), and finally output the fusion characteristics themselves, at different scales. The algorithm fully mines the spectral spatial information of images, fuses and generates deeper image features, and establishes more accurate relationships between characteristic values. With the deepening of the model's network depth—to prevent the emergence of network degradation and other problems—we use a residual network algorithm to fuse the semantic information between the different scale graph convolution and strengthen the connections between different residual network algorithms. This not only deepens the network depth, but also solves the problem of network degradation.

The main contributions of this paper are as follows:

1. Firstly, for HSI processing, a multi-scale convolution algorithm is introduced to mine the shallow features of the image from a multi-directional and multi-level perspective, which significantly enhances the model's feature-mining ability.
2. Secondly, a multi-scale graph convolution algorithm is proposed. The algorithm uses a different number of feature points to form different graphs and constructs the internal relationships of features at different scales so that the relationships between multiple feature points can be comprehensively constructed.
3. Lastly, a residual network algorithm is used to fuse the semantic information between different scale graph convolution, so as to further improve the quality of the deep features. By enhancing the relationship between different residual network algorithms, the problem of network degradation is effectively solved, and the classification accuracy improved.

The rest of the paper is organized as follows. In Section 2 we briefly review the most relevant literature on hyperspectral remote sensing image classification. In Section 3, a multi-scale feature-mining learning algorithm is proposed, in which each model component (multi-scale graph convolution, feature fusion and so on) is explained in detail. The effectiveness of proposed method will be evaluated in Section 4. Finally, we give the summarization in Section 5.

## 2 Related work

Since there is a rich literature on hyperspectral remote sensing image classification, we only review some most relevant work. The following is summarized from the current research status of multi-scale convolution neural networks, graph convolution neural networks, and residual networks.

**Multi-scale convolutional neural network:** The CNN is widely used in the field of remote sensing. Bai-Sen Liu *et al.* [7] proposed a multi-scale convolutional neural network model, which solved the problem of over-fitting based on a general CNN. The model solved the problem of single-input data scale, adjusting the neighbourhood of multiple pixels and classifying them to the same size. Mesut et al. [13] proposed AlexNet to extract the feature information of hyperspectral remote sensing images. ResNet based method[14] and DenseNet based method[15] are used to increase the feature representative ability of HSIs by fully using useful features. In order to extract more information, SAGP[18] and PRAN[16] use attention mechanism to get more complicated feature to represent the HSIs. However, the above methods did not fully utilize the spectral and spatial information of HSIs.

**Graph Convolutional Network:** The traditional convolution model has a disadvantage in that it can only deal with fixed square regions and the convolution kernel weight is fixed. However, a graph convolution network based on the traditional convolution algorithm shows good performance in processing non-square regions, although it cannot guarantee pixel relationships under the premise of limited iteration times. Siyuan Liu *et al.* [8] proposed a deep feature learning model based on label consistency (DFL-LC). Based on the HSI features extracted by a multi-scale convolutional neural network (MSCNN), the model used the graph convolution method to obtain the relationship between pixels, and the label consistency of single pixels (LCSP) and label consistency of group pixels (LCGP) to distinguish the features. This method could extract more representative and judicious image features, offering more quantitative and qualitative advantages. Danfeng Hong *et al.* [9] put forward a miniGCN model, which calculated large GCNs in a small batch way and combined a GCN and a CNN to break the performance bottleneck of a single model. This model not only solved the problem of irregular data representation and analysis, but also removed a major shortcoming of the traditional GCN model, that is, its significant computational overhead. However, in the field of HSI classification, the imbalance between high and limited labelled samples has always been a difficult problem to solve. To solve it, Anshu Sha *et al.* [10] proposed the edge-convolutional-graph convolutional-network model, which analysed the performance of the edge conditional convolution over graph and modelling it on the basis of the corresponding topological relationships.

**Residual network:** Kumari Pooja *et al.* [11] proposed a multi-scale expanded residual CNN (MDR-CNN) model. Based on a CNN, the model combined extended multi-scale convolution with the concept of residual connection. Moreover, the real-time data was classified while the acceptance domain was utilised. Because of its combination with the residual network, the model could learn better features than the coded ones. In addition, although HSIs have become increasingly important in the field of remote sensing, they will always be disturbed by noise. Consequently, many existing models have proposed de-noise models to solve this problem. Yuan Yuan et al. [12] proposed a new multi-scale residual learning network, in which image noise was obtained through a noise estimation network, eliminating noise inconsistency accordingly. Meanwhile, the multi-scale residual model they used further eliminated noise on the basis of the former.

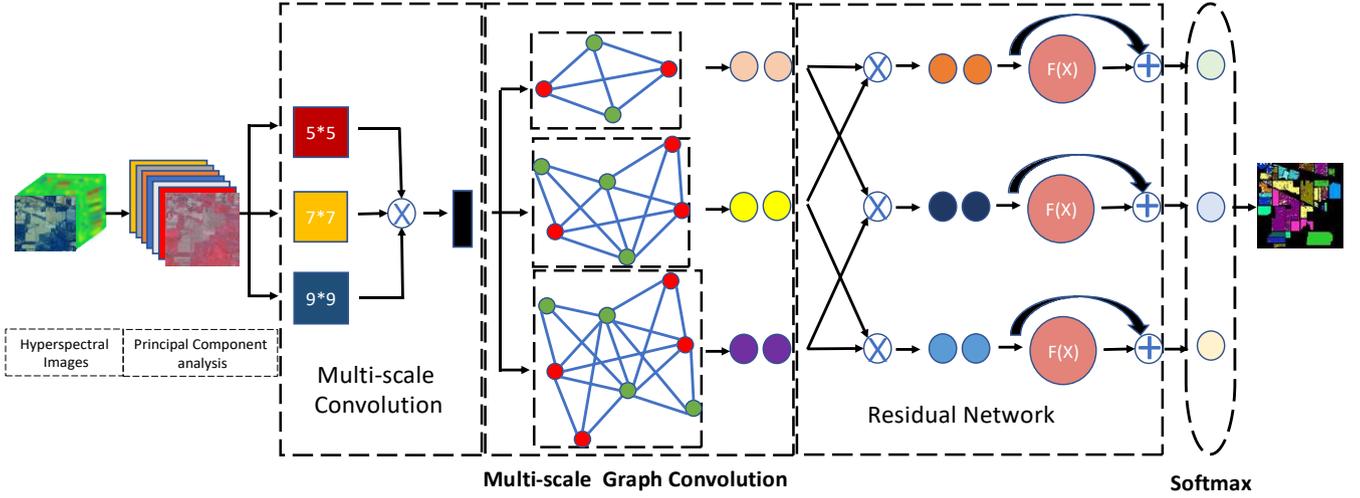

**Figure 1.** The MGRNet algorithm model.

## 3 Method: Proposed Framework

The multi-scale graph convolution combined with residual network fusion algorithm (MGRNet) proposed in this paper is shown in Figure 1. The basic principles are as follows: Firstly, the original HSI is processed using principal component analysis (PCA) to retain 99.99% of its semantic information and extract dimensionality reduction features. Then, the dimensionality reduction features are input into the multi-scale convolution algorithm for shallow feature mining. Next, the shallow features are input into the multi-scale graph convolution algorithm to build the relationships between the eigenvalues. Furthermore, the multi-scale information obtained by graph convolution is cross-fused and input into the residual network algorithm for deep feature mining. Finally, Softmax software is used to predict the final features and complete the recognition of hyperspectral remote sensing images. In the following sections of this section, we cover each aspect of the MGRNet algorithm in detail.

### 3.1 Multi-Scale algorithm model

Firstly, for feature extraction of HSIs, a multi-scale convolution algorithm is introduced in this paper, the core ideas of which are as follows: Using multiple convolution kernels to mine and extract image features at the same time, feature information of different scales can be obtained. Clearly, compared with existing single convolution kernel feature mining techniques, multi-convolution kernel simultaneous mining can obtain more HSI information, avoid the loss of key features or information, and significantly improve the ability of model mining. Specifically, the convolutional layer formula is as follows:

$$X_{i*j} = \sigma_i(\sum_{s=1}^{n_i-1} X_{i-1*s} \otimes W_{i*s*j} + b_{i*j}) \quad (1)$$

Where $X_{i-1*s}$ represents the *s-th* feature map of the *i-1* layer; $W_{i*s*j}$ is the convolution kernel weight between the *s-th* feature map of the *i-1* layer and the *j-th* feature map of the *i-1* layer; $b_{i*j}$ represents the bias of the *j-th* feature map of the *i-th* layer; $n_i - 1$ represents the number of feature maps contained in layer *i-1*; $\otimes$ represents the convolution operation; $\sigma_i$ represents the activation function of layer *i*, this model adopting the *ReLU* function as the activation function. A multi-scale convolution kernel is simultaneously used for shallow feature mining. The scale of the convolution kernel is represented by *k*. The formula of the multi-scale convolution is as follows:

$$X_{i*j}^{(k)} = \sigma_i(\sum_{s=1}^{n_i-1} X_{i-1*s} \otimes W_{i*s*j}^{(k)} + b_{i*j}^{(k)}) \quad (2)$$

### 3.2 Multi-Scale Graph Convolution

Traditional convolution computing plays an active role in feature mining, but it is impossible to extract the relationships between features. However, graph convolution not only retains deep features, but also constructs the relationships between them. Nevertheless, existing graph convolution methods generally only use a fixed graph in the process of node convolution, which cannot accurately reflect the relationships between pixels. In addition, the domain size of the graph is generally fixed, so the spectral spatial information of different local regions cannot be captured with any degree of flexibility. To solve these problems, a new algorithm for multi-scale graph convolution is proposed. By using a different number of feature points to form different graphs, we mine the multi-scale spectral spatial information of images, construct the internal relationships of features at different scales, and comprehensively construct the relationships between multiple feature points. The multi-scale graph convolution operation is shown in Figure 2.

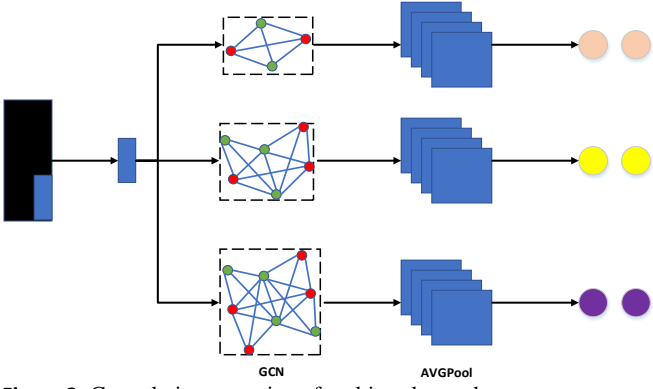

**Figure 2.** Convolution operation of multi-scale graph.

The formula for calculating graph convolution is as follows:

$$g_i = ReLU(\sum_{j \in N(i)} a_j f(j) + b_j) \quad (3)$$

where $N(i)$ represents the feature set of layer $i$; $a_j$ represents the weight coefficient of the $j$-th eigenvalue; $b_j$ represents the bias of the $j$-th eigenvalue; $g_i$ represents the characteristic output of layer $i$. Based on traditional graph convolution, the convolution of different scale graphs is constructed to more comprehensively establish the relationships between features. Formula (3) is rewritten as follows with the scale of the figure represented by $k$:

$$g_i^k = ReLU(\sum_{j \in N(i)} a_j^k f(j) + b_j^k) \quad (4)$$

This module establishes the relationships between different feature points through graph convolution of different scales. However, the graph comprises many feature points containing numerous and complicated relationships between features, and the relationships between feature points constructed are not sufficiently close, which affects classification accuracy. To solve this problem, a pool layer is added after the graph convolution layer, and the feature of the feature graph obtained by graph convolution is deleted and reselected. On the one hand, pooling reduces the size of the feature graph, making the relationships between the feature values clearer, and improving the correlation between feature points. On the other hand, average pooling reduces the generated feature map to the same scale, which is conducive to subsequent feature fusion. The pooling layer formula is as follows:

$$S_i = Pool_{Average}(g_i) \quad (5)$$

where $Pool_{Average}(*)$ represents the average value of returned feature points within a certain size region.

### 3.3 Residual Network

Compared with a traditional convolutional network, a deep network has more network layers and a more complex structure, so it exhibits stronger feature learning and feature expression abilities. However, with the deepening of the network, network degradation becomes a serious problem, the accuracy of classification and recognition decreasing rather than increasing. To solve these problems, firstly, the pairwise fusion of the different relationships between feature points is generated by convolution of a multi-scale graph using a concatenate function. Then, the fusion features are input into the residual network algorithm to build residual networks with different features for deeper feature mining. Furthermore, by merging the characteristic graphs of different graph convolution, different residual network algorithms are realised. We then use the residual network algorithm to fuse the semantic information between different scale graph convolution and make full use of the mined multi-scale information to extract more expressive deep features. The residual network model is shown in Figure 3.

The formula of feature cross fusion is as follows:

$$Out_1 = concatenate(Integ_1 \oplus Integ_2)$$
$$Out_2 = concatenate(Integ_1 \oplus Integ_3) \quad (6)$$
$$Out_3 = concatenate(Integ_2 \oplus Integ_3)$$

Among them, $Integ_1$, $Integ_2$, and $Integ_3$ represent the feature sequences generated by the convolution of the following figures of three different scales. $Out_1$, $Out_2$, and $Out_3$ represent the pairwise fusion of feature outputs of the graphs of three different scales. $concatenate(*)$ represents the feature fusion function; $\oplus$ stands for fusion operation. The residual module formula is as follows:

$$y = F(x, \{w_i\}) + w_s x \quad (7)$$

where $x$ represents the input value of the residual module; $y$ is the output result of the residual module; $F(x, \{w_i\})$ is the residual mapping, $\{w_i\}$ is the ownership value of the residual module; $w_s$ is the weight matrix, and $x$ is projected linearly to give it the same dimension as the residual mapping.

### 3.4 Feature Fusion

Different residual network mining techniques generate different deep features with more expressive abilities. To make full use of the mined feature information, the concatenate function is used to fuse the deep feature sequences integrated by different residual networks to generate the final fusion features. The fusion algorithm formula is as follows:

$$Output = concatenate(feature_1 \oplus feature_2 \oplus feature_3)$$
$$(8)$$

where $feature_1$ represents the feature sequence generated by the residual network of fused feature $Out_1$; $feature_2$ represents the feature sequences generated by the residual network of fused feature $Out_2$; $feature_3$ represents the feature sequences generated by the residual network of fused feature $Out_3$; $Output$ represents final f-usion feature output. Finally, Softmax is used to predict the final fusion features.

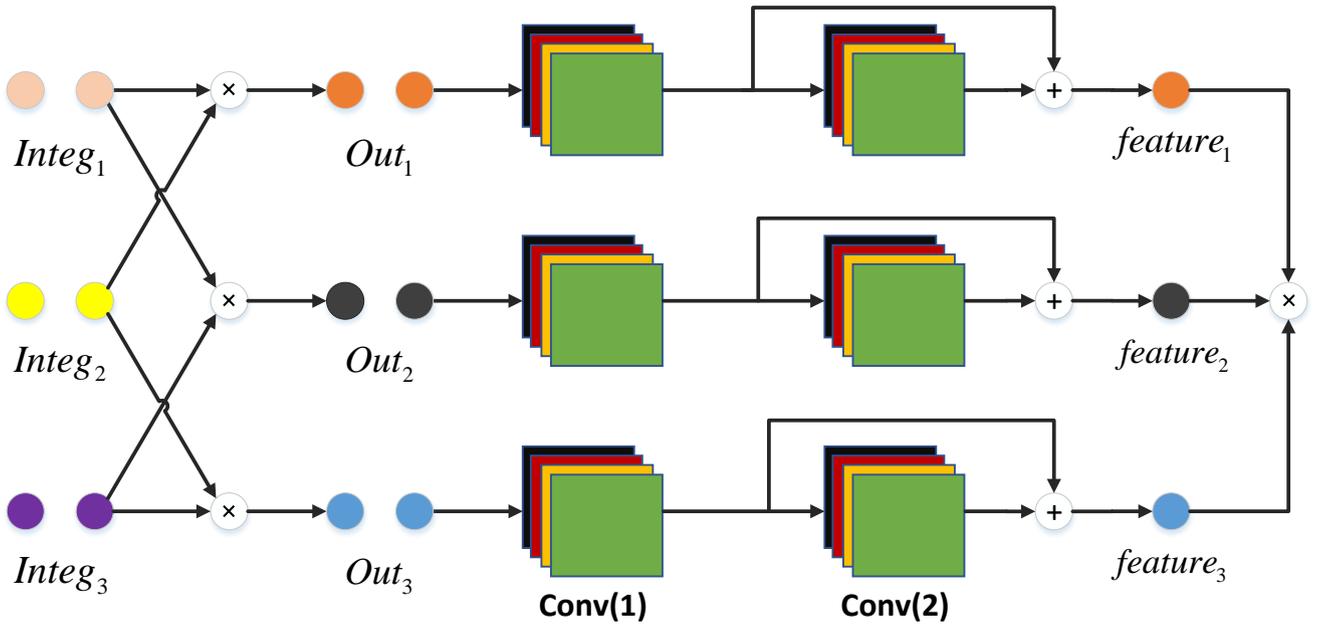

**Figure 3.** Residual network model structure.

## 4 Results and Discussions

In this paper, experiments were conducted on three open HSI datasets: the Indian Pines, Pavia University, and Salinas Valley datasets. The performance of the model was evaluated using three indexes—that is, the overall accuracy (OA), average accuracy (AA), and kappa coefficient. To verify the classification ability of the proposed MGRNET algorithm, it was compared with other HSI classification algorithms, including ALEXNET[13], DENSENET[15], FSSFNET[17], RESNET[14], SAGP[18], PRAN[16], AML[4], and DFL-LC[8].

### 4.1 Data Sets

The first data set was the Indian Pines dataset captured by an Aviris sensor. This dataset has a spatial resolution of 20 m/pixel and an image size of $145 \times 145$ pixels. The original data contained 224 bands, and 200 effective bands were retained after processing, covering 16 crop categories. The categories, number of samples, and representative colours of this dataset are shown in Table 1. PCA dimension reduction processing was carried out on the dataset images to reduce the dimension to 100-D.

The second data set was the Pavia University dataset collected by a ROSIS sensor. This dataset has a spatial resolution of 1.3 m/pixel and an image size of $610 \times 340$ pixels. A total of 115 bands were obtained by the sensor, and 103 bands were retained after processing. There were nine types of ground features in total. The categories, number of samples, and representative colours of this dataset are shown in Table 2. PCA dimension reduction processing was carried out on the dataset images to reduce the dimension to 50-D.

The third data set was the Salinas Valley dataset collected by an Aviris sensor. This dataset has a spatial resolution of 3.7 m/pixel and an image size of $512 \times 217$ pixels. The original data contained 224 bands, and 204 effective bands were retained after removing the bands with severe water vapour absorption. The dataset included 16 crop categories. Each class, number of samples, and representative colour of this dataset are shown in Table 3. PCA dimension reduction processing was carried out on the dataset images to reduce the dimension to 29-D.

**Table 1.** The Indian Pines dataset class and number of samples.

| No. | Class | Samples | Colour |
|---|---|---|---|
| 1 | Alfalfa | 46 | |
| 2 | Corn | 237 | |
| 3 | Corn-notill | 1428 | |
| 4 | Grass-pasture | 483 | |
| 5 | Corn-mintill | 830 | |
| 6 | Grass-trees | 730 | |
| 7 | Grass-pasture-mowed | 28 | |
| 8 | Hay-windrowed | 478 | |
| 9 | Oats | 20 | |
| 10 | Soybean-notill | 972 | |
| 11 | Soybean-mintill | 2455 | |
| 12 | Soybean-clean | 593 | |
| 13 | Wheat | 205 | |
| 14 | Woods | 1265 | |
| 15 | Buildings-Grass-Trees-Drives | 386 | |
| 16 | Stone-Steel-Towers | 93 | |
| | Total | 10249 | |

**Table 2.** The Pavia University dataset class and number of samples.

| No. | Class | Samples | Colour |
|---|---|---|---|
| 1 | Asphalt | 6631 | |
| 2 | Meadows | 18649 | |
| 3 | Gravel | 2099 | |
| 4 | Trees | 3064 | |

| 5 | Painted metal sheets | 1345 | |
| 6 | Bare Soil | 5029 | |
| 7 | Bitumen | 1330 | |
| 8 | Self-Blocking Bricks | 3682 | |
| 9 | Shadows | 947 | |
| | Total | 42776 | |

Table 3. The Salinas Valley dataset class and number of samples.

| No. | Class | Samples | Colour |
| --- | --- | --- | --- |
| 1 | Brocoli-green-weeds-1 | 2009 | |
| 2 | Brocoli-green-weeds-2 | 3726 | |
| 3 | Fallow | 1976 | |
| 4 | Fallow-rough-plow | 1394 | |
| 5 | Fallow-smooth | 2678 | |
| 6 | Stubble | 3959 | |
| 7 | Celery | 3579 | |
| 8 | Grapes-untrained | 11271 | |
| 9 | Soil-vinyard-develop | 6203 | |
| 10 | Corn-senesced-green-weeds | 3278 | |
| 11 | Lettuce-romaine-4wk | 1068 | |
| 12 | Lettuce-romaine-5wk | 1927 | |
| 13 | Lettuce-romaine-6wk | 916 | |
| 14 | Lettuce-romaine-7wk | 1070 | |
| 15 | Vinyard-untrained | 7268 | |
| 16 | Vinyard-vertical-trellis | 1807 | |
| | Total | 54129 | |

**4.2 Experimental Results**

Experiments were carried out on three datasets based on the above methods. Among them, 20% of the samples of all models in the Indian Pines dataset were selected as the training set, 10% of the samples of all models in the Pavia University and Salinas Valley datasets were selected as the training set, and the remaining samples were selected as the test set to evaluate the classification performance. Studying the influence of the proportion of training sets on each model, the superiority of the algorithm proposed in this paper could be further demonstrated. We changed the proportions of the training set, conducted experiments on the three datasets, and observed the changes of the performance indexes of each algorithm when the number of training samples was reduced. Among them, the proportion of training sets for all models on the Indian Pines datasets was reduced to 10% and 5% respectively, and the proportion of training sets for all models on the Pavia University and Salinas Valley datasets was reduced to 5% and 1%, respectively. The results of different models on different datasets are shown in Table 4, Table 5, Table 6, and Figure 4, Figure 5, and Figure 6, where the maximum value of each indicator is marked in bold.

It can be seen from the results that the three evaluation indexes of OA, AA, and Kappa coefficients obtained by our proposed MGRNet algorithm on three different datasets all reached higher values than those of the other eight comparison models. Among them, the MGRNET algorithm exhibited the most obvious classification performance advantage on the Indian Pines dataset. Specifically, the OA index was 3–13% higher than that of the other models, the AA index was 6–12% higher than that of the other models, and the Kappa coefficient was 8–15% higher than that of the other models. In the three datasets, the Indian Pines dataset had significantly fewer samples than the other two datasets and belonged to the small sample dataset. The results showed that the MGRNET algorithm exhibited better performance in small sample classification.

The classification diagram of remote sensing images obtained using different algorithms on the Indian Pines dataset is shown in Figure 4 (the proportion of samples used in the training sets was 20%). The classification accuracy of the MGRNET algorithm was higher and the noise was lower, achieving the optimal outcome.

Both the Pavia University and Salinas Valley datasets had a relatively large number of samples, and each algorithm generated a relatively high evaluation index. In this case, the classification performance of the MGRNET algorithm was still better—that is, compared to the comparison models, the various evaluation indexes improved by 0.5–9%, proving the superiority of the MGRNET algorithm. The FSSFNET algorithm also exhibited good classification performance, indicating that when the number of samples was sufficient, the spatial feature relationship between pixels became more important. The classification diagrams of remote sensing images obtained by different algorithms on the Pavia University and Salinas Valley datasets are shown in Figure 5 and Figure 6, respectively (the proportion of samples used in the training sets was 20%). The results can prove that multi-scale image convolution combined with a residual network algorithm plays a positive role in remote sensing image classification.

From the results of changing the proportion of samples used in the training sets, a decrease in the proportion of samples used in the training sets reduced the performance index of each algorithm. However, even if the proportion of samples used in the training sets decreased incrementally, the MGRNet algorithm still maintained its good classification performance, its performance indicators still being higher than the other models. As can be seen from the table, with a decrease in the proportion of samples used in the training sets, the classification performance of the MGRNet algorithm decreased relatively less. This suggests that the MGRNet algorithm was less affected by the proportion of samples used in the training sets and exhibited better stability. For the Pavia University dataset, when the proportion of samples used in the training set was 5%, the result of the PRAN algorithm OA was the highest, suggesting that better results could be obtained by increasing the attention mechanism in irregular samples. For the Salinas Valley dataset, when the proportion of samples used in the training set was 5%, the result of the FSSFNet algorithm OA was the highest, suggesting that with morphological rule samples, the spatial feature relationship of pixels plays an important role. Overall, the experimental results proved that the MGRNet algorithm proposed in this paper exhibited better feature learning abilities and could maintain its good classification performance in case with fewer training samples.

**Table 4.** OA, AA, and Kappa coefficients obtained by different algorithms on the Indian Pines dataset.

| Methods | OA (20%) | AA (20%) | Kappa (20%) | OA (10%) | AA (10%) | Kappa (10%) | OA (5%) | AA (5%) | Kappa (5%) |
|---|---|---|---|---|---|---|---|---|---|
| AlexNet | 0.8152 | 0.8283 | 0.8038 | 0.7361 | 0.7232 | 0.6836 | 0.7046 | 0.6964 | 0.6510 |
| DenseNet | 0.7819 | 0.7774 | 0.7450 | 0.7051 | 0.6796 | 0.6287 | 0.6634 | 0.6482 | 0.5956 |
| FSSFNet | 0.7610 | 0.8370 | 0.8126 | 0.5955 | 0.7079 | 0.6605 | 0.6493 | 0.7391 | 0.7030 |
| PRAN | 0.8451 | 0.8313 | 0.8081 | 0.6402 | 0.6867 | 0.6380 | 0.7369 | 0.7519 | 0.7174 |
| ResNet | 0.8478 | 0.8273 | 0.8021 | 0.7556 | 0.7471 | 0.7101 | 0.6859 | 0.6801 | 0.6311 |
| SAGP | 0.8752 | 0.8407 | 0.8178 | 0.7566 | 0.7250 | 0.684 | 0.7333 | 0.6675 | 0.6183 |
| AML | 0.8844 | 0.8903 | 0.8748 | 0.8197 | 0.8213 | 0.8019 | 0.7704 | 07584 | 0.7216 |
| DFL-LC | 0.8902 | 0.8485 | 0.8264 | 0.8145 | 0.8129 | 0.7986 | 0.7640 | 0.7562 | 0.7186 |
| MGRNet | **0.9121** | **0.9043** | **0.8911** | **0.8260** | **0.8320** | **0.8077** | **0.7875** | **0.7602** | **0.7226** |

**Table 5.** OA, AA and Kappa coefficients obtained by different algorithms on the Pavia University dataset.

| Methods | OA (10%) | AA (10%) | Kappa (10%) | OA (5%) | AA (5%) | Kappa (5%) | OA (1%) | AA (1%) | Kappa (1%) |
|---|---|---|---|---|---|---|---|---|---|
| AlexNet | 0.9304 | 0.942 | 0.9231 | 0.9040 | 0.9258 | 0.9013 | 0.8403 | 0.8769 | 0.8365 |
| DenseNet | 0.9283 | 0.9348 | 0.9138 | 0.9098 | 0.9230 | 0.8977 | 0.8116 | 0.8192 | 0.7566 |
| FSSFNet | 0.9358 | 0.9507 | 0.9345 | 0.9161 | 0.9332 | 0.9112 | 0.8674 | 0.8706 | 0.8254 |
| PRAN | 0.9361 | 0.9475 | 0.9302 | **0.9294** | 0.9361 | 0.9147 | 0.8783 | 0.8922 | 0.8564 |
| ResNet | 0.9332 | 0.9457 | 0.9279 | 0.9247 | 0.9356 | 0.9144 | 0.8366 | 0.8594 | 0.8114 |
| SAGP | 0.9325 | 0.9329 | 0.9101 | 0.9084 | 0.9087 | 0.8777 | 0.8302 | 0.8224 | 0.7578 |
| AML | 0.9373 | 0.9444 | 0.9265 | 0.9225 | 0.9316 | 0.9237 | 0.8781 | 0.8961 | 0.8723 |
| DFL-LC | 0.9396 | 0.9420 | 0.9230 | 0.9258 | 0.9388 | 0.9224 | 0.8759 | 0.9010 | 0.8736 |
| MGRNet | **0.9447** | **0.9542** | **0.9394** | 0.9282 | **0.9445** | **0.9263** | **0.8824** | **0.9185** | **0.8921** |

**Table 6.** OA, AA and Kappa coefficients obtained by different algorithms on the Salinas Valley dataset.

| Methods | OA (10%) | AA (10%) | Kappa (10%) | OA (5%) | AA (5%) | Kappa (5%) | OA (1%) | AA (1%) | Kappa (1%) |
|---|---|---|---|---|---|---|---|---|---|
| AlexNet | 0.9758 | 0.9524 | 0.6179 | 0.9680 | 0.9406 | 0.6133 | 0.9281 | 0.9001 | 0.6009 |
| DenseNet | 0.9682 | 0.9398 | 0.9329 | 0.8689 | 0.8845 | 0.8712 | 0.8270 | 0.8381 | 0.8194 |
| FSSFNet | 0.9776 | 0.9578 | 0.9530 | **0.9722** | 0.9414 | 0.9347 | 0.9444 | 0.8992 | 0.8879 |
| PRAN | 0.8941 | 0.9169 | 0.9072 | 0.8604 | 0.8710 | 0.8567 | 0.7516 | 0.7929 | 0.7691 |
| ResNet | 0.9768 | 0.9529 | 0.9476 | 0.9609 | 0.9355 | 0.9282 | 0.9128 | 0.8863 | 0.8736 |
| SAGP | 0.9608 | 0.9291 | 0.9210 | 0.9569 | 0.8996 | 0.8884 | 0.9419 | 0.8844 | 0.8714 |
| AML | 0.9733 | 0.9548 | 0.9497 | 0.9554 | 0.9315 | 0.9252 | 0.9363 | 0.8858 | 0.8815 |
| DFL-LC | 0.9800 | 0.9599 | 0.9553 | 0.9653 | 0.9437 | 0.9258 | 0.9436 | 0.8972 | 0.8854 |
| MGRNet | **0.9817** | **0.9616** | **0.9572** | 0.9700 | **0.9457** | **0.9395** | **0.9505** | **0.9063** | **0.8953** |

### 4.3 Influence of Number of Iterations

To more accurately study the classification performance of the MGRNET algorithm, we changed the number of iterations and conducted experiments on the three datasets to observe the changes in each performance index of the MGRNET algorithm when the number of iterations increased. We set the number of iterations to be 20, 50, 100, 150, 200, 250, 300, 350, 400, 450, and 500 and evaluated the classification performance of each algorithm using OA, AA and Kappa coefficients. The results are shown in Figure 7, Figure 8, and Figure 9, below.

It can be seen from the experimental results that, with increasing iterations, the performance indicators of the MGRNET algorithm gradually increased. When the number of iterations increased to approximately 250, the performance indexes of the MGRNET algorithm tended to stabilise. After a certain number of iterations, the various parameters of our model tended to be optimal, the algorithm having a relatively stable classification performance. At the same time, the results also demonstrated that 500 iterations were sufficient, producing convincing results.

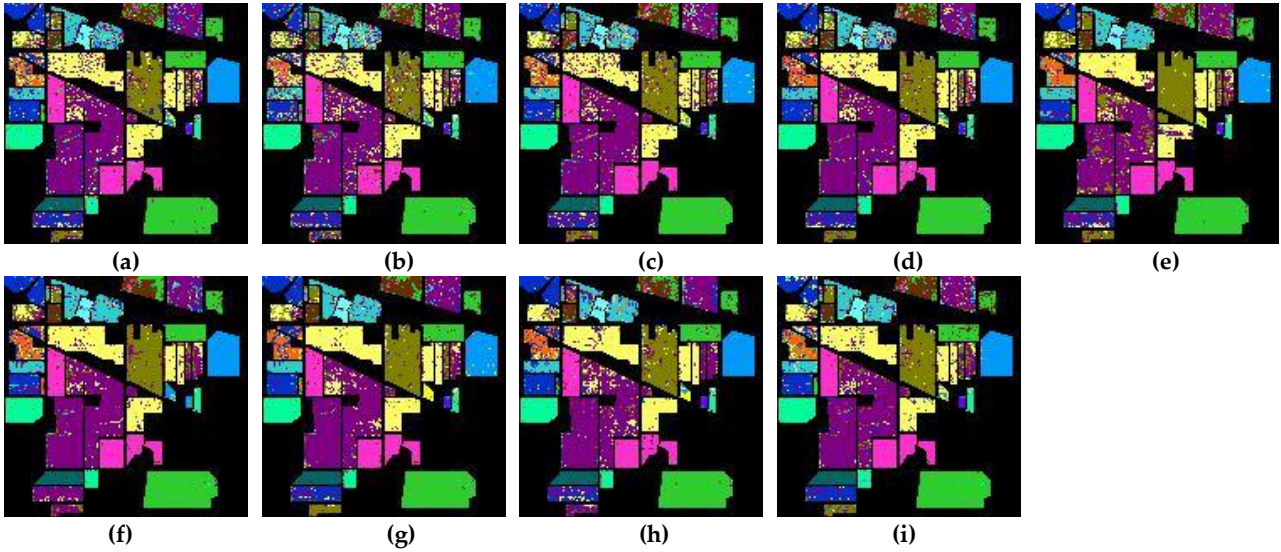

**Figure 4.** Classification of remote sensing images obtained using different algorithms on the Indian Pines dataset. (a) AlexNet, (b) DenseNet, (c) ResNet, (d) SAGP, (e) PRAN, (f) FSSFNet, (g) AML, (h) DFL-LC and (i) MGRNet.

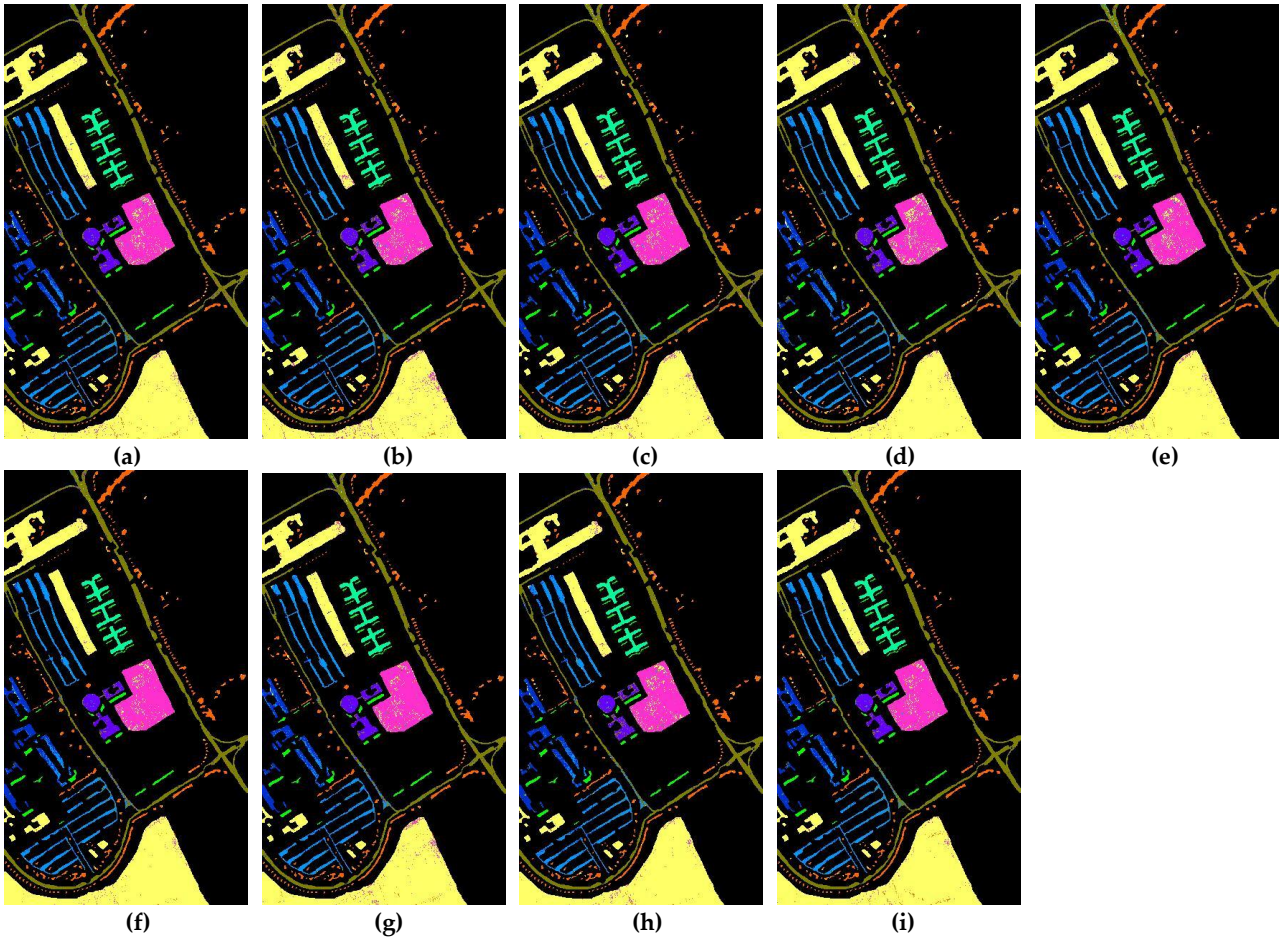

**Figure 5.** Classification of remote sensing images obtained using different algorithms on the Pavia University dataset. (a) AlexNet, (b) DenseNet, (c) ResNet, (d) SAGP, (e) PRAN, (f) FSSFNet, (g) AML, (h) DFL-LC and (i) MGRNet.

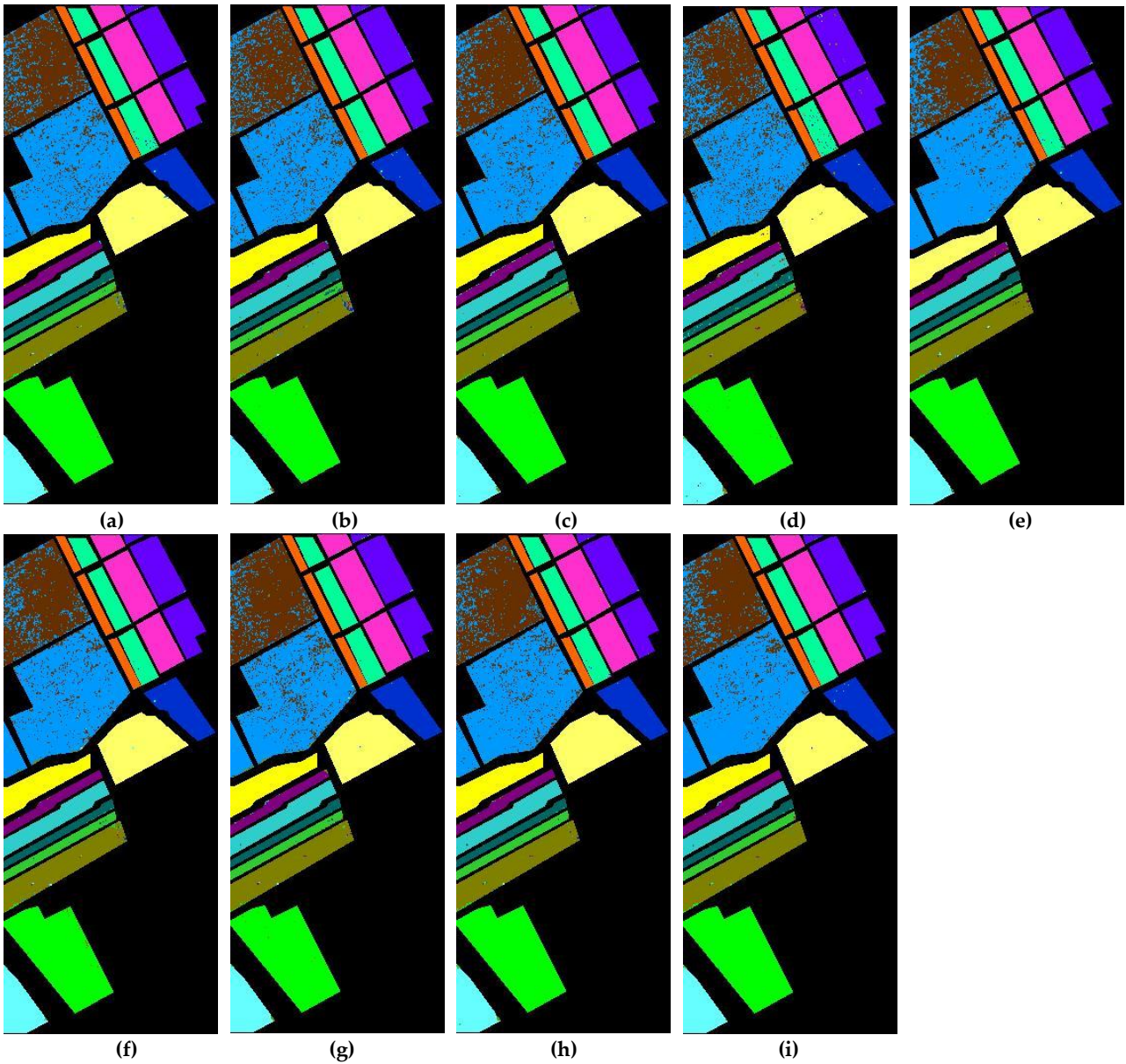

**Figure 6.** Classification of remote sensing images obtained using different algorithms on the Salinas Valley dataset. (a) AlexNet, (b) DenseNet, (c) ResNet, (d) SAGP, (e) PRAN, (f) FSSFNet, (g) AML, (h) DFL-LC and (i) MGRNet.

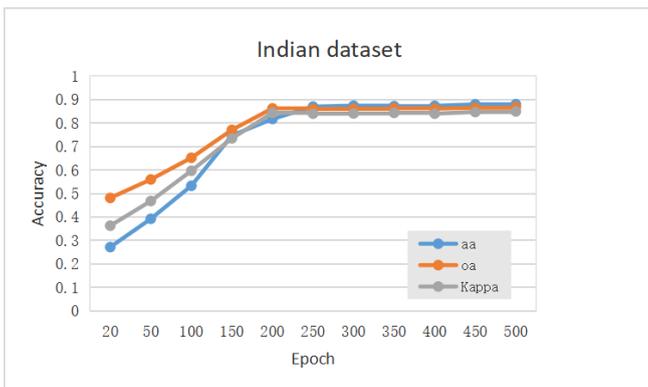

**Figure 7.** OA, AA, and Kappa coefficients on the Indian Pines dataset with different iterations.

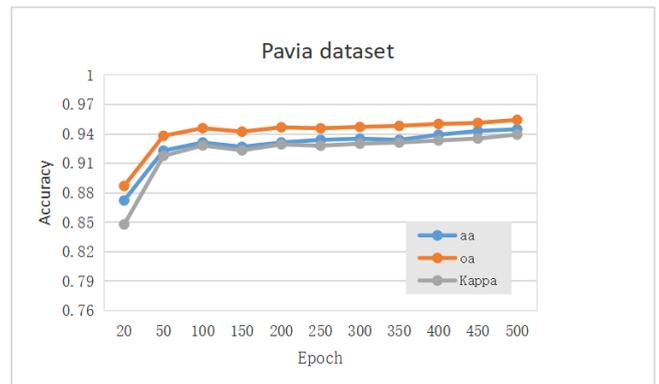

**Figure 8.** OA, AA and Kappa coefficients obtained from the Pavia University dataset with different iteration times.

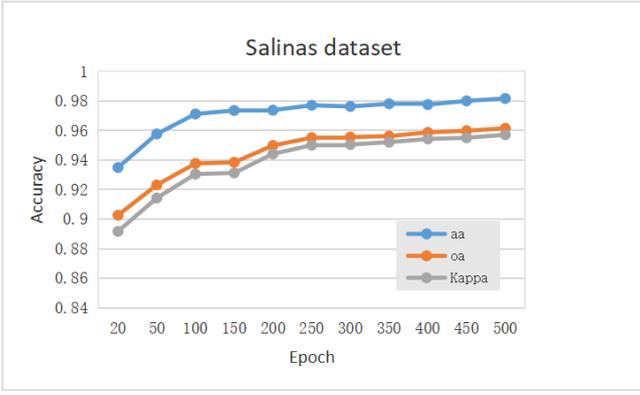

**Figure 9.** OA, AA and Kappa coefficients obtained on the Salinas Valley dataset with different iterations.

### 4.4. Comparison Between Traditional Graph Convolution and Multi-scale Graph Convolution

To verify the effectiveness of the multi-scale graph convolution proposed in this paper, we used traditional graph convolution and multi-scale graph convolution to conduct experiments on the three datasets. The traditional graph convolution used 16, 36, and 64 eigenvalues, respectively, to construct the graph convolution at different scales, while the multi-scale graph convolution simultaneously used the above three scales to construct the multi-scale graph convolution set. The traditional graph convolution used 20% of the samples from all models in the Indian Pines dataset as the training set, 10% of the samples from all models in the Pavia University and Salinas Valley datasets as the training set, and the remaining samples as the test set to evaluate the classification performance. The results are shown in Table 7, where the maximum value of each index is marked in bold.

It can be seen from the experimental results that the evaluation indexes of the multi-scale graph convolution on the three datasets were all higher than those of the traditional graph convolution. Consequently, the multi-scale graph convolution proposed in this paper played a positive role in improving the classification performance.

### 4.5 Experimental Results of Different Sub-modules

To study the role of sub-modules in remote sensing image classification, we used different sub-modules to conduct experiments on the three datasets. The sub-modules were respectively set to be no multi-scale convolution module (NC), no multi-scale graph convolution module (NG), and no residual network module (NR). 20% of the samples from all models in the Indian Pines dataset were used as the training set, 10% of the samples from all models in the Pavia University and Salinas Valley datasets were used as the training set, and the remaining samples were used as the test set to evaluate the classification performance. The results are shown in Table 8, where the highest value of each indicator is marked in bold.

From the results, the three sub-modules all show good classification performance. In the Indian Pines dataset, the sample still maintained a high dimension after PCA processing, which suggests that feature mining is very important, and that the multi-scale convolution and residual network modules will have a greater impact. It can be seen that when the multi-scale convolution or residual network module was removed, the classification accuracy decreased significantly. The classification results of sub-modules of the Pavia University and Salinas Valley datasets were similar, the evaluation indexes being the highest when the three sub-modules were used together. The MGRNET algorithm achieved the best results in the three groups of tests, which proved that the proposed algorithm played a positive role in remote sensing image classification.

## 5 CONCLUSION

In this paper, we proposed a multi-scale graph convolution combined with residual network cross-fusion algorithm (MGRNet) to fully mine and make full use of image features. Firstly, the MGRNet algorithm uses multi-scale convolution to mine the shallow features of the image, before using a different number of eigenvalues to construct different feature graphs to establish the relationships between features at diff-rent scales. Next, the output of different graph convolution is fused, and the fused features input into different residual networks for deep feature mining. Finally, the depth feature sequences integrated by different residual networks are fused. The algorithm achieved ideal results on three open datasets, proving the proposed method to be feasible.

**Table 7.** OA, AA and Kappa coefficients obtained by traditional graph convolution and multi-scale graph convolution on different datasets.

| Methods | Indian Pines | | | Pavia University | | | Salinas Valley | | |
|---|---|---|---|---|---|---|---|---|---|
| | OA | AA | Kappa | OA | AA | Kappa | OA | AA | Kappa |
| MGRNet_G64 | 0.8657 | 0.8838 | 0.8676 | 0.9237 | 0.9440 | 0.9257 | 0.9761 | 0.9552 | 0.9501 |
| MGRNet_G36 | 0.8921 | 0.8671 | 0.8481 | 0.9365 | 0.9484 | 0.9315 | 0.9805 | 0.9601 | 0.9555 |
| MGRNet_G16 | 0.8591 | 0.8645 | 0.8451 | 0.9302 | 0.9440 | 0.9257 | 0.9788 | 0.9581 | 0.9534 |
| MGRNet | **0.9121** | **0.9043** | **0.8911** | **0.9447** | **0.9543** | **0.9394** | **0.9817** | **0.9615** | **0.9572** |

**Table 8.** OA, AA and Kappa coefficients obtained by different sub-modules on different datasets.

| Methods | Indian Pines | | | Pavia University | | | Salinas Valley | | |
|---|---|---|---|---|---|---|---|---|---|
| | OA | AA | Kappa | OA | AA | Kappa | OA | AA | Kappa |
| MGRNet_NC | 0.8681 | 0.8673 | 0.8489 | 0.9309 | 0.9467 | 0.9292 | 0.9785 | 0.9560 | 0.9511 |
| MGRNet_NG | 0.9020 | 0.8923 | 0.8771 | 0.9308 | 0.9473 | 0.9302 | 0.9774 | 0.9545 | 0.9494 |
| MGRNet_NR | 0.8648 | 0.8794 | 0.8624 | 0.9359 | 0.9485 | 0.9316 | 0.9766 | 0.9549 | 0.9498 |
| MGRNet | **0.9121** | **0.9043** | **0.8911** | **0.9447** | **0.9543** | **0.9393** | **0.9817** | **0.9616** | **0.9572** |

## Acknowledgment

This work was supported partly by Chongqing Research Program of Basic Research and Frontier Technology under Grant cstc2019jcyjmsxmX0571.